\documentclass[10pt,twocolumn,letterpaper]{article}

\usepackage{cvpr}
\usepackage{times}
\usepackage{epsfig}
\usepackage{graphicx}
\usepackage{amsmath}
\usepackage{amssymb}
\usepackage{amsmath}
\usepackage{amssymb}
\usepackage{algorithm}
\usepackage{algorithmicx}
\usepackage{algpseudocode}
\usepackage{graphics}
\usepackage{threeparttable}
\usepackage{color}
\usepackage[normalem]{ulem}
\usepackage{multirow}
\usepackage{float}
\usepackage{amsfonts}
\usepackage{bm}
\usepackage{array}

\makeatletter
\newcommand{\thickhline}{%
    \noalign {\ifnum 0=`}\fi \hrule height 1pt
    \futurelet \reserved@a \@xhline
}
\makeatother

\cvprfinalcopy 

\def\cvprPaperID{2628} 

\ifcvprfinal\fi
\begin{document}

\title{A Causal And-Or Graph Model for Visibility Fluent Reasoning in \\Tracking Interacting Objects}

\author{Yuanlu Xu$^{1}$\thanks{Yuanlu Xu and Lei Qin contributed equally to this paper. This work is supported by ONR MURI Project N00014-16-1-2007, DARPA XAI Award N66001-17-2-4029, and NSF IIS 1423305, 1657600, and in part by National Natural Science Foundation of China: 61572465, 61390510, 61732007. The correspondence author is Xiaobai Liu.}\,\,,\; Lei Qin$^{2\,*}$,\; Xiaobai Liu$^3$,\; Jianwen Xie$^4$,\; Song-Chun Zhu$^1$\\
$^1$University of California, Los Angeles\quad $^2$Inst. Computing Technology, Chinese Academy of Sciences\\
$^3$Dept. Computer Science, San Diego State University \quad
$^4$Hikvision Research Institute, USA\\{\tt\small yuanluxu@cs.ucla.edu, qinlei@ict.ac.cn, xiaobai.liu@mail.sdsu.edu}\\
{\tt\small jianwen.xie@hikvision.com, sczhu@stat.ucla.edu}
}

\maketitle

\begin{abstract}

Tracking humans that are interacting with the other subjects or environment remains unsolved in visual tracking, because the visibility of the human of interests in videos is unknown and might vary over time. In particular, it is still difficult for state-of-the-art human trackers to recover complete human trajectories in crowded scenes with frequent human interactions.
In this work, we consider the visibility status of a subject as a fluent variable, whose change is mostly attributed to the subject's interaction with the surrounding, e.g., crossing behind another object, entering a building, or getting into a vehicle, etc. We introduce a Causal And-Or Graph (C-AOG) to represent the causal-effect relations between an object's visibility fluent and its activities, and develop a probabilistic graph model to jointly reason the visibility fluent change (e.g., from visible to invisible) and track humans in videos. We formulate this joint task as an iterative search of a feasible causal graph structure that enables fast search algorithm, e.g., dynamic programming method. We apply the proposed method on challenging video sequences to evaluate its capabilities of estimating visibility fluent changes of subjects and tracking subjects of interests over time. Results with comparisons demonstrate that our method outperforms the alternative trackers and can recover complete trajectories of humans in complicated scenarios with frequent human interactions.

\end{abstract}

\maketitle

\vspace{-2mm}
\section{Introduction}
\vspace{-1mm}

\begin{figure}[ptb]
\centering
\includegraphics[width=\linewidth]{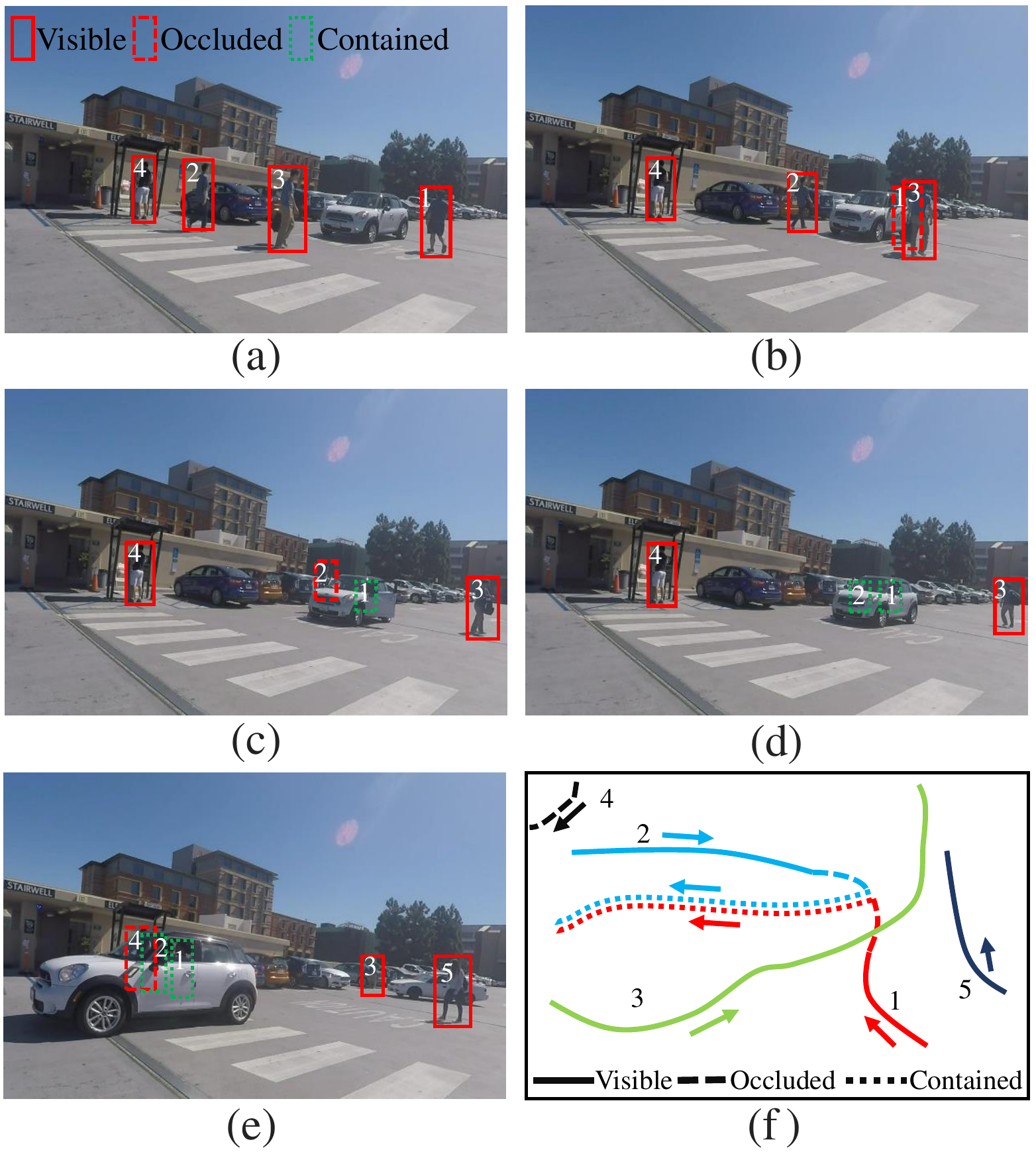}
\caption{\textbf{Illustration of visibility fluent changes.} There are three states: visible, occluded, contained. When a person approaches a vehicle, its state changes from ``visible'' to ``occluded'' to ``contained'', such as the person$_1$ and person$_2$ (a-e). When a vehicle passes, the person$_4$ is occluded. The state of person$_4$ changes from ``visible'' to ``occluded'' in (d-e). (f) shows the corresponding top-view trajectories of different persons. The numbers are the persons' IDs. The arrows indicate the moving direction.}
\label{fig:intro}
\vspace{-10pt}
\end{figure}

Tracking objects of interest in videos is a fundamental computer vision problem that has great potentials in many video-based applications, e.g., security surveillance, disaster response, and border patrol. In these applications, a critical problem is how to obtain the complete trajectory of the object of interest while observing it moving in the scene through camera view. This is a challenging problem since an object of interest might undergo frequent interactions with the surrounding, e.g., entering a vehicle or a building, or with the other objects, e.g., passing behind another subject. With these interactions, the visibility status of a subject will be varying over time, e.g., changing from ``invisible'' to ``visible'' and vice versa. In the literature, most state-of-the-art trackers utilize appearance or motion cues to localize subjects in video sequences and are likely to fail to track the subjects whose visibility status keep changing.

To deal with the above challenges, in this work, we propose to explicitly reason subjects' visibility status over time, while tracking the subjects of interests in surveillance videos. Traditional trackers are likely to fail when the target become invisible due to occlusion, our proposed method could jointly infer objects' locations and visibility fluent changes, thus helping to recover the complete trajectories. The proposed techniques, with slight modifications, can be generalized to other scenarios, e.g., hand-held cameras, driverless vehicles, etc.

The key idea of our method is to introduce a fluent variable for each subject of interest to explicitly indicate his/her visibility status in videos. Fluent was firstly used by Newton to denote the time varying status of an object. It is also used to represent the varying object status in commonsense reasoning~\cite{mueller2014commonsense}. In this paper, the visibility status of objects can be described as fluents varying over time. As illustrated in Fig.~\ref{fig:intro}, the person$_3$ and person$_5$ are walking through the parking lot, while the person$_1$ and person$_2$ are entering a sedan. The visibility status of person$_1$'s and person$_2$'s changes first from ``visible'' to ``occluded'', and then to ``contained''. This group example demonstrates how objects' visibility fluents change over time along with their interactions to the surrounding.

We introduce a graphical model, i.e. Causal And-Or graph (C-AOG), to represent the causal relationships between object's activities (actions/sub-events) and object's visibility fluent changes. The visibility status of an object might be caused by multiple actions, and we need to reason the actual causality from videos. These actions are alternative choices that lead to the same occlusion status, and form the Or-nodes. Each leaf node indicates an action or sub-event that can be described by And-nodes. Taking the videos shown in Fig.~\ref{fig:intro} for instance, the status of ``occluded'' can be caused by the following actions: (i) walking behind a vehicle; (ii) walking behind a person; or (iii) inertial action that maintains the fluent unchanged.

The basic hypothesis of this model is that, for a particular scenario (e.g., parking-lot), there are only a limited number of actions that can cause the fluent to change. Given a video sequence, we need to create the optimal C-AOG and select the best choice for each Or-node in order to obtain the optimal causal parse graph, which is shown as red lines in Fig.~\ref{fig:caog}(a).

We develop a probabilistic graph model to reason object's visibility fluent changes using C-AOG representation. Our formula integrates object tracking purposes as well to enable joint solution of tracking and fluent change reasoning, which are mutually beneficial. In particular, for each subject of interest, our method uses two variables to represent (i) subjects' positions in videos; and (ii) visibility status as well as the best causal parse graph. We utilize a Markov Chain Prior model to describe the transitions of these variables, i.e., the current state of a subject is only dependent on the previous state. We then reformulate the problem into an Integer Linear Programming model, and utilize dynamic programming to search the optimal states over time.

In experimental evaluations, the proposed method is tested on a set of challenging sequences that include frequent human-vehicle or human-human intersections. Results show that our method can readily predict the correct visibility status and recover the complete trajectories. In contrast, most of the alternative trackers can only recover part of the trajectories due to the occlusion or containment.

\textbf{Contributions}. There are three major contributions of the proposed framework: (i) a Causal And-Or Graph (C-AOG) model to represent object visibility fluents varying over time; (ii) a joint probabilistic formulation for object tracking and fluent reasoning; and (iii) a new occlusion reasoning dataset to cover objects with diverse fluent changes.

\vspace{-2mm}
\section{Related Work} \label{sec:literature}
\vspace{-1mm}

The proposed research is closely related to the following three research streams in computer vision and AI.

\textbf{Multiple object tracking} has been extensively studied in the past decades. In the past literatures, tracking-by-detection has become the mainstream framework~\cite{UndirHyperGraphTracker,IDNetFlowTracker,xu2016multi,xu2017multi,dong2017occlusion,dongxingping2018}. Specifically, a general detector~\cite{FelzenszwalbCVPR2010,ren2015faster} is first applied to generate detection proposals, and then data association techniques~\cite{berclaz2011multiple,GMMCPTracker,SPATrackerCVPR16} are employed to link detection proposals over time in order to get object trajectories. Our approach also follows this pipeline, but is more focused on the reasoning of object visibility status.

\textbf{Tracking interacting objects} studies a more specific problem of tracking entangled objects. Some works~\cite{TrackInteract14,TrackInteract16,ModelInteract16} try to model the object appearing and disappearing phenomena globally, yielding strong assumptions on appearance, location or motion cues. On the contrary, other works attempt to model human-object and human-human interactions under specific scenarios, such as social activities~\cite{group12,group17}, team sports~\cite{sport13}, and people carrying luggage~\cite{luggage13}. In this paper, we propose a more principled way to track objects with both short-term interactions, e.g., passing behind another object, or long-term interactions, e.g., entering a vehicle and moving together.

\textbf{Causal-effect reasoning} is a popular topic in AI but has not received much attentions in the field of computer vision. It studies, for instances, the difference between co-occurrence and causality, and aims to learn causal knowledge automatically from low-level observations, e.g., images or videos. There are two popular causality models: Bayesian Network~\cite{Griffiths2005,Pearl2009} and grammar models~\cite{Griffiths2007,WeiContain16}. Grammar models~\cite{xu2013reid,fang2018_3dpose,wang2018} are powerful tool for modeling high-level human knowledge in specific domains. Notably, Fire and Zhu~\cite{AmyCausal16} have introduced a causal grammar to infer causal-effect relationship between object's status, e.g., door open/close, and agent's actions, e.g., pushing the door. They studied this problem using manually designed rules and video sequences in lab settings. In this work, we extend the causal grammar models to infer objects' visibility fluent and ground the task on challenging videos in surveillance systems.

\begin{figure}[ptb]
\centering
\includegraphics[width=\linewidth]{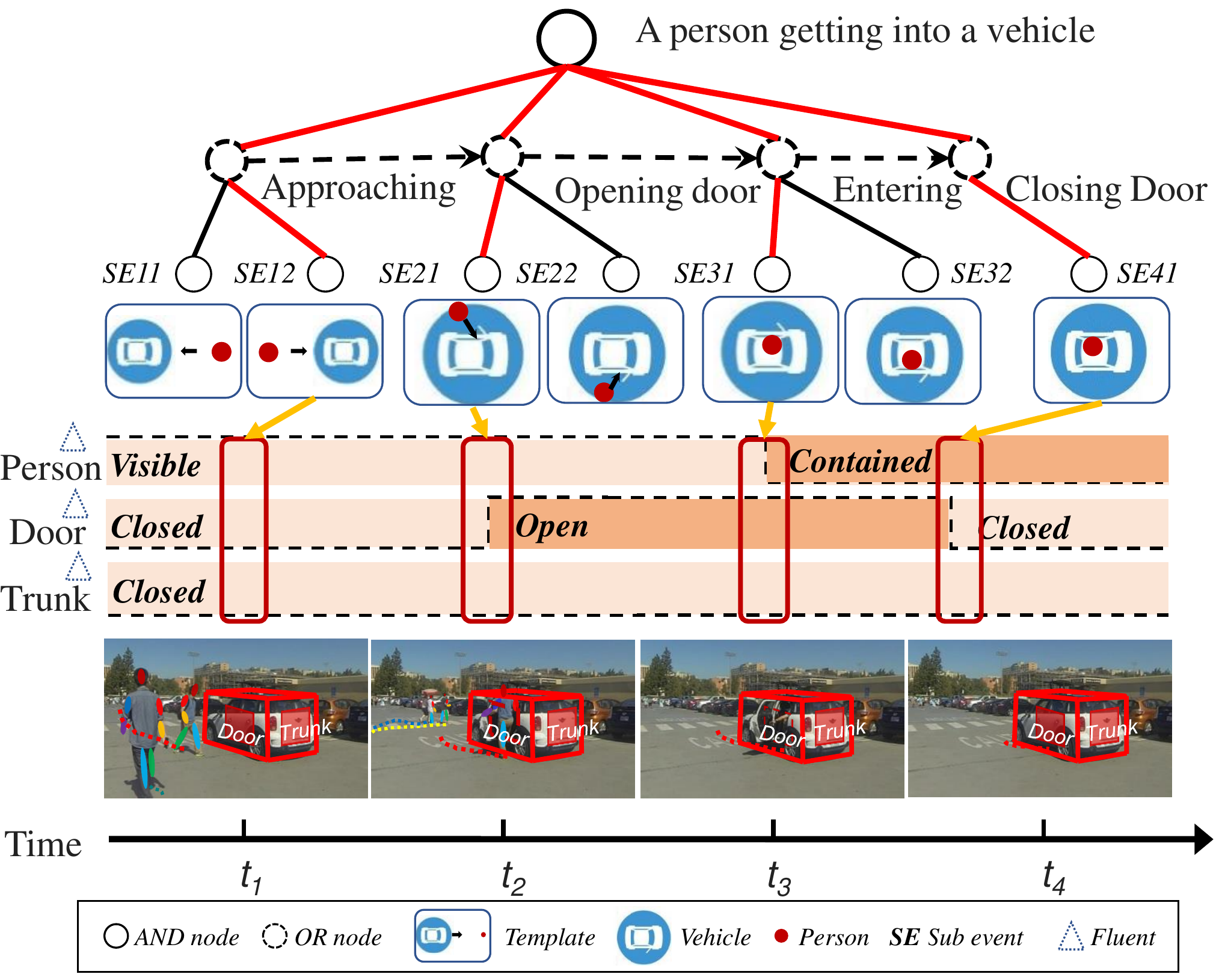}
\caption{\textbf{Illustration of a person's actions and her visibility fluent changes} when she enters a vehicle.}
\label{fig:pg}
\vspace{-10pt}
\end{figure}

\vspace{-2mm}
\section{Representation} \label{sec:model}
\vspace{-1mm}

\begin{figure*}[ptb]
\centering
\includegraphics[width=\linewidth]{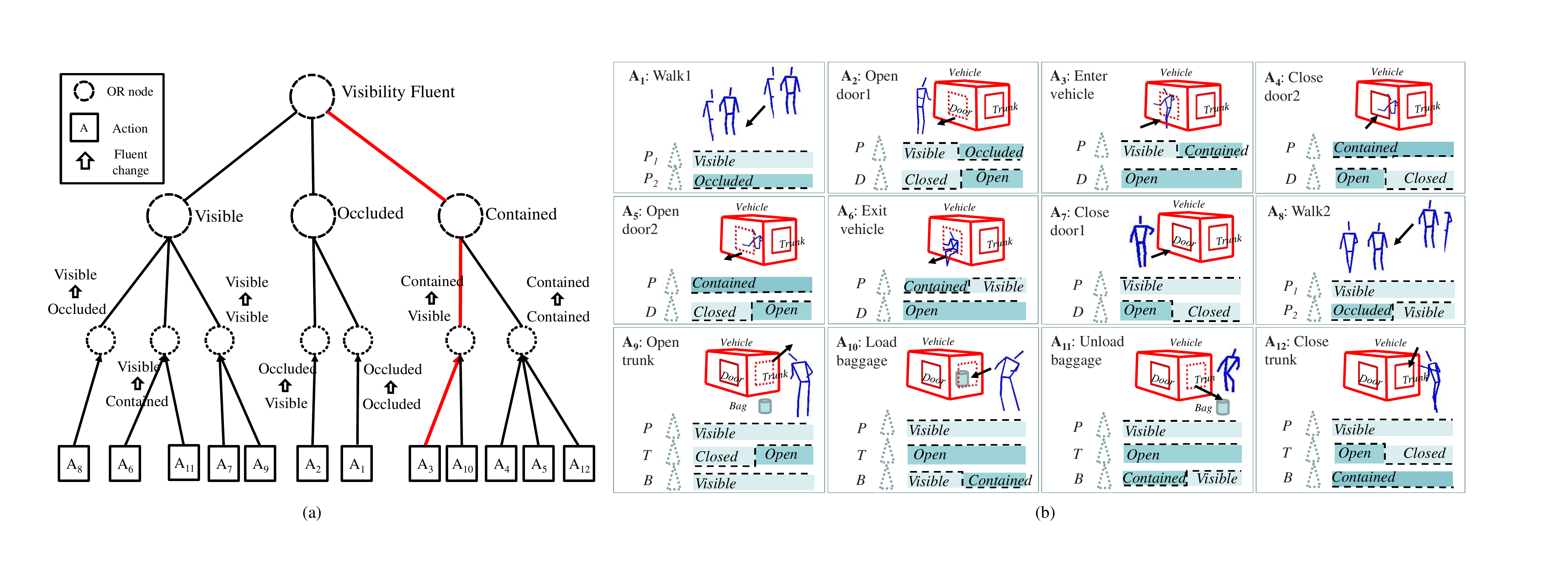}
\caption{\textbf{(a) The proposed Causal And-Or Graph (C-AOG) model for the fluent of visibility.} We use a C-AOG to represent the visibility status of an subject. Each OR node indicates a possible choice and an arrow shows how visibility fluent transits among states. \textbf{(b) A series of atomic actions that could possibly cause visibility fluent change.} Each atomic action describes interactions among people and interacting objects. ``P'', ``D'', ``T'', ``B'' denotes ``person'', ``door'', ``trunk'', ``bag'', respectively. The dash triangle denotes fluent. The corresponding fluent could be ``visible'', ``occluded'' or ``contained'' for a person; ``open'', ``closed'' or ``occluded'' for a vehicle door or truck. See text for more details.}
\label{fig:caog}
\vspace{-10pt}
\end{figure*}

In this paper, we define three states for visibility fluent reasoning: \textbf{visible}, (partially/fully) \textbf{occluded}, and \textbf{contained}. Most multiple object tracking methods are based on tracking-by-detection framework, which obtain good performance in visible and partially occluded situations. However, when full occlusions take place, these trackers usually regard the disappearing-and-reappearing objects as new  objects. Although objects in fully occluded and contained states are invisible, there are still evidences to infer the locations of objects and fill-in the complete trajectory. We can distinguish object being fully occluded and object being contained from three empirical observations.

Firstly, \emph{motion independence}.
In fully occluded state, such as a person staying behind a pillar, the motion of the person is independent of the pillar. While in contained state, such as a person sitting in a vehicle, or a bag in the trunk, the position and motion of the person/bag would be the same as  the vehicle. Therefore, the inference of the visibility fluent of the object is important in tracking objects accurately in a complex environment.

Secondly, \emph{coupling actions and object fluent changes}.
For example, as illustrated in Fig.~\ref{fig:pg}, if a person gets into a vehicle, the related sequential atomic actions are: approaching a vehicle, opening the vehicle door, getting into the vehicle, and closing the vehicle door; the related object fluent changes are vehicle door {\em closed $\rightarrow$ open $\rightarrow$ closed}. The fluent change is a consequence of agent actions. If the fluent-changing actions do not happen, the object should maintain its current fluent. For example, a person that is contained in a vehicle will remain contained unless he/she opens the vehicle door and gets out of the vehicle.

Thirdly, \emph{visibility in the alternative camera views}.
In full occlusion state, such as a person occluded by a pillar, though the person could not be observed from the current viewpoint, he/she could be seen from the other viewpoints; while in contained state, such as a person in a vehicle, this person could not be seen from any viewpoints.

In this work, we mainly study the interactions of humans and the developed methods can also be expanded to other objects, e.g., animals.

\vspace{-2mm}
\subsection{Causal And-Or Graph}
\vspace{-1mm}

In this paper, we propose a Causal And-Or Graph (C-AOG) to represent the action-fluent relationship, as illustrated in Fig.~\ref{fig:caog}(a). A C-AOG has two types of nodes: (i) Or-nodes that represent the variations or choices, and (ii) And-nodes that represent the decompositions of the top-level entities. The arrows indicate the causal relations between actions and fluent transitions. For example, a C-AOG can be used to expressively model a series of action-fluent relations.

The C-AOG is capable of representing multiple alternatives for causes of occlusion and potential transitions. There are four levels in our C-AOG: visibility fluents, possible states, state transitions and agent actions. Or nodes represent alternative causes in visibility fluents and state levels; that is, one fluent can have multiple states  and one state can have multiple transitions. An event can be decomposed into several atomic actions and represented by an And-node, e.g., an event of a person getting into a vehicle is a composition of four atomic actions: approaching the vehicle, opening the door, entering the vehicle, and closing the door.

Given a video sequence $I$ with length $T$ and camera calibration parameters $H$, we represent the scene $\mathcal{R}$ as
\begin{equation} \begin{aligned}
	& \mathcal{R}=\left\lbrace O_t:\, t=1,2,...,T \right\rbrace, \\
	& O_t = \left\lbrace o_{t}^{i}:\, i=1,2,...,N_t \right\rbrace,
\end{aligned} \end{equation} 	
where $O_t$ denotes all the objects at time $t$, and $N_t$ is the size of $O_t$, i.e., the number of objects at time $t$. $N_t$ is unknown and will be inferred from observations.
Each object $o_{t}^{i}$ is represented with its location $l_t^i$ (i.e., bounding boxes in the image) and appearance features $\phi_{t}^{i}$. To study the visibility fluent of a subject, we further incorporate a state variable $s_{t}^{i}$ and an action label $a_{t}^{i}$, that is,
\begin{equation}
	o_{t}^{i} = \left(l_{t}^{i},\, \phi_{t}^{i},\, s_{t}^{i}, a_{t}^{i} \right).
\end{equation}
Thus, the state of a subject is defined as
\begin{equation}
	s_{t}^{i} \in S = \left\lbrace \text{\,visible, occluded, contained\,} \right\rbrace.
\end{equation}
We define a series of atomic actions $\Omega = \{a_i: i=1,\dots,N_a\}$ that might change the visibility status, e.g., walking, opening vehicle door, etc. Fig.~\ref{fig:caog}(b) illustrates a small set of actions $\Omega$ covering the most common interactions among people and vehicles.

Our goal is to jointly find subject locations in video frames and estimate their visibility fluents $M$ from the video sequence $I$. Formally, we have
\begin{equation}\begin{aligned}
    & M = \{pg_t: t = 1,2,\dots,T\}, \\
    & pg_t = \{o_{t}^{i} = (l_{t}^{i}, \phi_{t}^{i}, s_{t}^{i}, a_{t}^{i})\,|\,i=1,2,...,N_t\},
\end{aligned} \end{equation}
where $pg_t$ can be determined by the optimal causal parse graph at time $t$.

\vspace{-2mm}
\section{Problem Formulation} \label{sec:formula}
\vspace{-1mm}

According to Bayes' rule, we can solve our joint object tracking and fluent reasoning problem by maximizing a posterior (MAP),
\begin{equation} \label{equation:score} \begin{aligned}
     M^* &\,=\, \underset{M}{\arg\,\max}\; p(M|I; \theta) \\
         &\,\varpropto\, \underset{M}{\arg\,\max}\, p(I|M; \theta) \cdot p(M; \theta) \\
         &\,=\, \underset{M}{\arg\,\max}\; \frac{1}{Z} \exp\,\{-\mathcal{E}(M;\theta)- \mathcal{E}(I|M;\theta)\}.
\end{aligned}\end{equation}
The {\bf prior} term $\mathcal{E}(M;\theta)$ measures the temporal consistency between successive parse graphs. Assuming $G$ is a Markov Chain structure, we can decompose $\mathcal{E}(M;\theta)$ as
\begin{equation} \begin{aligned} \label{equation:tr_score}
  	\mathcal{E}(M;\theta) &= \sum_{t=1}^{T-1} \mathcal{E}(pg_{t+1}|pg_t) \\
   	&= \sum_{t=1}^{T-1} \sum_{i=1}^{N_t} \Phi(l_{t+1}^{i},l_{t}^{i},s_{t}^{i})+ \Psi(s_{t+1}^{i},s_{t}^{i}, a_{t}^{i}).
\end{aligned} \end{equation}

The first term $\Phi(\cdot)$ measures the location displacement. It calculates the transition distance between two successive frames and is defined as:
\begin{equation}\small
	\Phi(l_{t+1}^{i},l_{t}^{i},s_{t}^{i}) =
	\begin{cases}
	\delta(\mathcal{D}_s(l_{t+1}^{i},l_{t}^{i}) > \tau_s),\; s_{t}^{i} = \text{Visible},\\
	1,\quad\quad\quad\quad\quad\quad\quad\quad\; s_{t}^{i} = \text{Occ, Con},
	\end{cases}
\end{equation}
where $\mathcal{D}_s(\cdot,\cdot)$ is the Euclidean distance between two locations on the 3D ground plane, $\tau_s$ is the speed threshold and $\delta(\cdot)$ is an indicator function. The location displacement term measures the motion consistency of object in successive frames.

The second term $\Psi(\cdot)$ measures the state transition energy and is defined as:
\begin{equation} \label{equation:state}
    \Psi(s_{t+1}^{i},s_{t}^{i},a_{t}^{i}) = -\log p(s_{t+1}^{i}|s_{t}^{i},a_{t}^{i}),
\end{equation}
where $p(s_{t+1}^{i}|s_{t}^{i},a_{t}^{i})$ is the action-state transition probability, which can be learned from the training data.

The {\bf likelihood} term $\mathcal{E}(I|M;\theta)$ measures how well each parse graph explains the data, which can be decomposed as
\begin{equation} \begin{aligned}
    \mathcal{E}(I|M;\theta) &= \sum_{t=1}^{T} \mathcal{E}(I_t|pg_t) \\
    &= \sum_{t=1}^{T} \sum_{i=1}^{N_t}\Upsilon(l_{t}^{i},\phi_{t}^{i},s_{t}^{i}) + \Gamma(l_{t}^{i},\phi_{t}^{i},a_{t}^{i}),
\end{aligned} \end{equation}
where $\Upsilon(\cdot)$ measures the likelihood between data and object fluents, and $\Gamma(\cdot)$ measures the likelihood between data and object actions. Given each object $o_{t}^{i}$, the energy function $\Upsilon(\cdot)$ is defined as:
\begin{equation} \label{equation:image_evidence}
  \Upsilon(l_{t}^{i},\phi_{t}^{i},s_{t}^{i}) =
  \begin{cases}
  1 - h_o(l_{t}^{i},\phi_{t}^{i}), & s_{t}^{i} = \text{Visible}, \\
  \sigma(\mathcal{D}_\varsigma(\varsigma_{1}^{i},\varsigma_{2}^{i})), & s_{t}^{i} = \text{Occluded},\\
  1 - h_c(l_{t}^{i},\phi_{t}^{i}), & s_{t}^{i} = \text{Contained},
  \end{cases}
\end{equation}
\begin{figure}[ptb]
\centering
\includegraphics[width=0.8\linewidth]{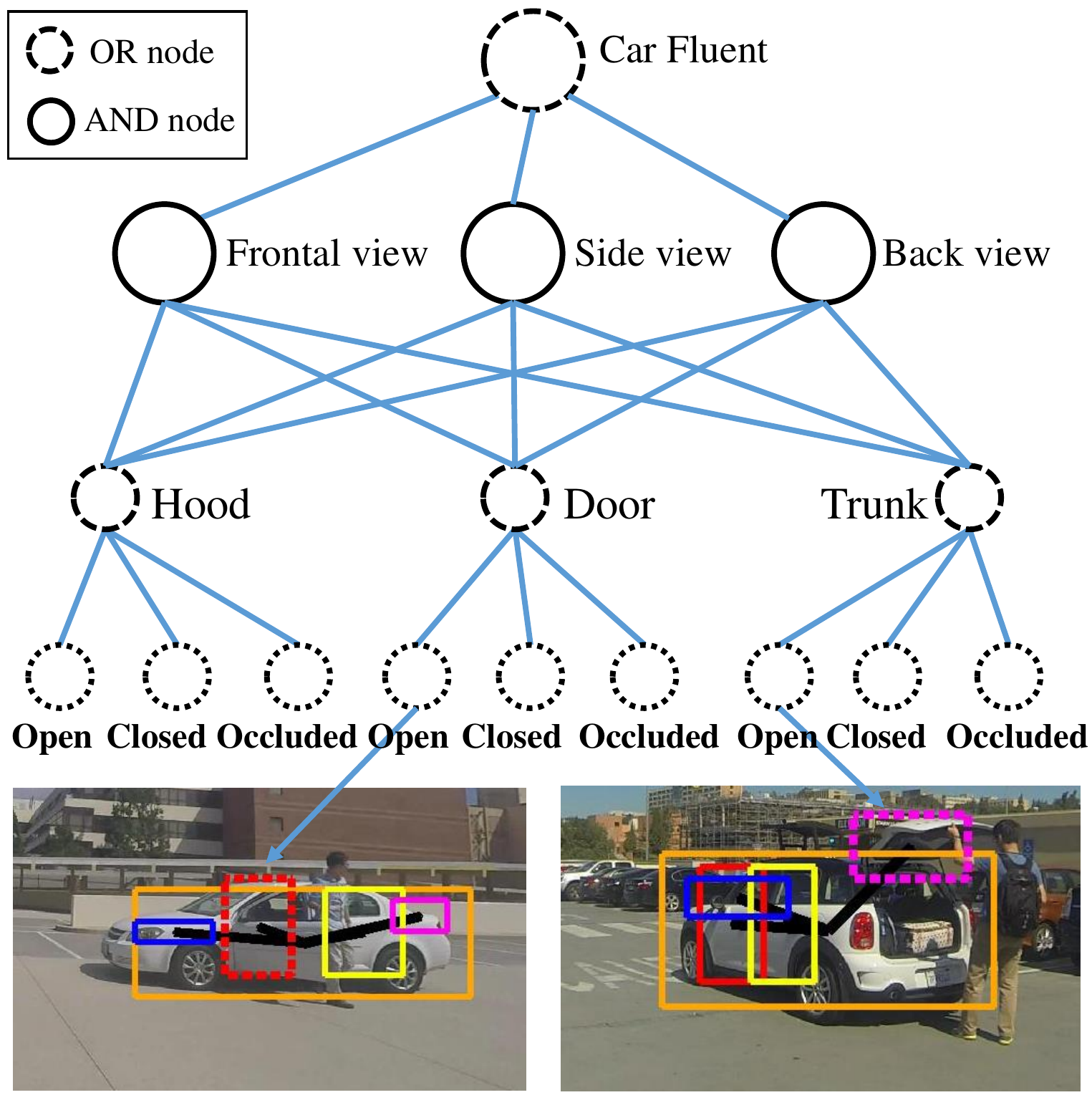}
\caption{\textbf{Illustration of Hierarchical And-Or Graph.} The vehicle is decomposed into different views, semantic parts and fluents. Some detection results are drawn below, with different colored bounding boxes denoting different vehicle parts, solid/dashed boxes denoting state ``closed''/``open''.}
\label{fig:CarFluent}
\vspace{-10pt}
\end{figure}
where $h_o(\cdot)$ indicates the object detection score, $h_c(\cdot)$ indicates the container (i.e., vehicles) detection score, and $\sigma(\cdot)$ is the sigmoid function. When an object is in either visible or contained state, appearance information can describe the probability of the existence of itself or the object containing it (i.e., container) at this location. When an object is occluded, there is no visual evidence to determine its state. Therefore, we utilize temporal information to generate candidate locations. We employ the SSP algorithm~\cite{pirsiavash2011globally} to generate trajectory fragments (i.e., tracklets). The candidate locations are identified as misses in complete object trajectories. The energy is thus defined as the cost of generating a virtual trajectory at this location. We compute this energy by computing the visual discrepancy between a neighboring tracklet $\varsigma_{1}^{i}$ before this moment and a neighboring tracklet $\varsigma_{2}^{i}$ after this moment. The appearance descriptor of a tracklet is computed as the average pooling of image descriptor over time. If the distance is below a threshold $\tau_\varsigma$, a virtual path is generated to connect these two tracklets using B-spline fitting.

The term $\Gamma(l_{t}^{i},\phi_{t}^{i},a_{t}^{i})$ is defined over the object actions observed from data. In this work, we study the fluents of human and vehicles, that is,
\begin{equation} \label{equation:fluent}
    \Gamma(l_{t}^{i},\phi_{t}^{i},a_{t}^{i})= \sigma(\mathcal{D}_h(l_{t}^{i},\phi_{t}^{i}|a_{t}^{i})) + \sigma(\mathcal{D}_v(l_{t}^{i},\phi_{t}^{i}|a_{t}^{i})),
\end{equation}
where $\sigma(\cdot)$ is the sigmoid function.  The definitions of the two data-likelihood terms $\mathcal{D}_h$ and $\mathcal{D}_v$ are introduced in the rest of this section.

A \textbf{human} is represented by his/her skeleton, which consists of multiple joints estimated by sequential prediction technology~\cite{wei2016convolutional}. The feature of each joint is defined as the relative distances of this joint to four saddle points(two shoulders, the center of the body, and the middle between the two hipbones). The relative distances are normalized by dividing the length of head to eliminate the influence of scale. A feature vector $\omega_t^h$ concatenating the features of all joints is extracted, which is assumed to follow a Gaussian distribution:
\begin{equation} \label{equation:pose}
    \mathcal{D}_h(l_{t}^{i},\phi_{t}^{i}|a_{t}^{i}) = -\log\, N(\omega_t^h;\mu_{a_{t}^{i}},\Sigma_{a_{t}^{i}}),
\end{equation}
where $\mu_{a_{t}^{i}}$ and $\Sigma_{a_{t}^{i}}$ are the mean and the covariance of the action $a_{t}^{i}$ respectively, which are obtained from the training data.

A \textbf{vehicle} is described with its viewpoint, semantic vehicle parts, and vehicle part fluents. The vehicle fluent is represented by a Hierarchical And-Or Graph, as illustrated in Fig.~\ref{fig:CarFluent}.
The feature vector of vehicle fluent $\omega^v$ is obtained by computing fluent scores on each vehicle part and concatenating them together.
We compute the average pooling feature $\varpi_{a_i}$ for each action $a_i$ over the training data as the vehicle fluent template. Given vehicle fluent $\omega^v_t$ computed on image $I_t$, the distance $\mathcal{D}_v(l_t^i,\phi_t^i|a_{t}^{i})$ is defined as
\begin{equation} \label{equation:carfluent}
    \mathcal{D}_v(l_t^i,\phi_t^i|a_{t}^{i}) = \|\,\omega^v_t - \varpi_{a_{t}^{i}}\,\|_2.
\end{equation}

\vspace{-2mm}
\section{Inference}\label{infer}
\vspace{-1mm}

\begin{figure}[ptb]
\centering
\includegraphics[width=0.85\linewidth]{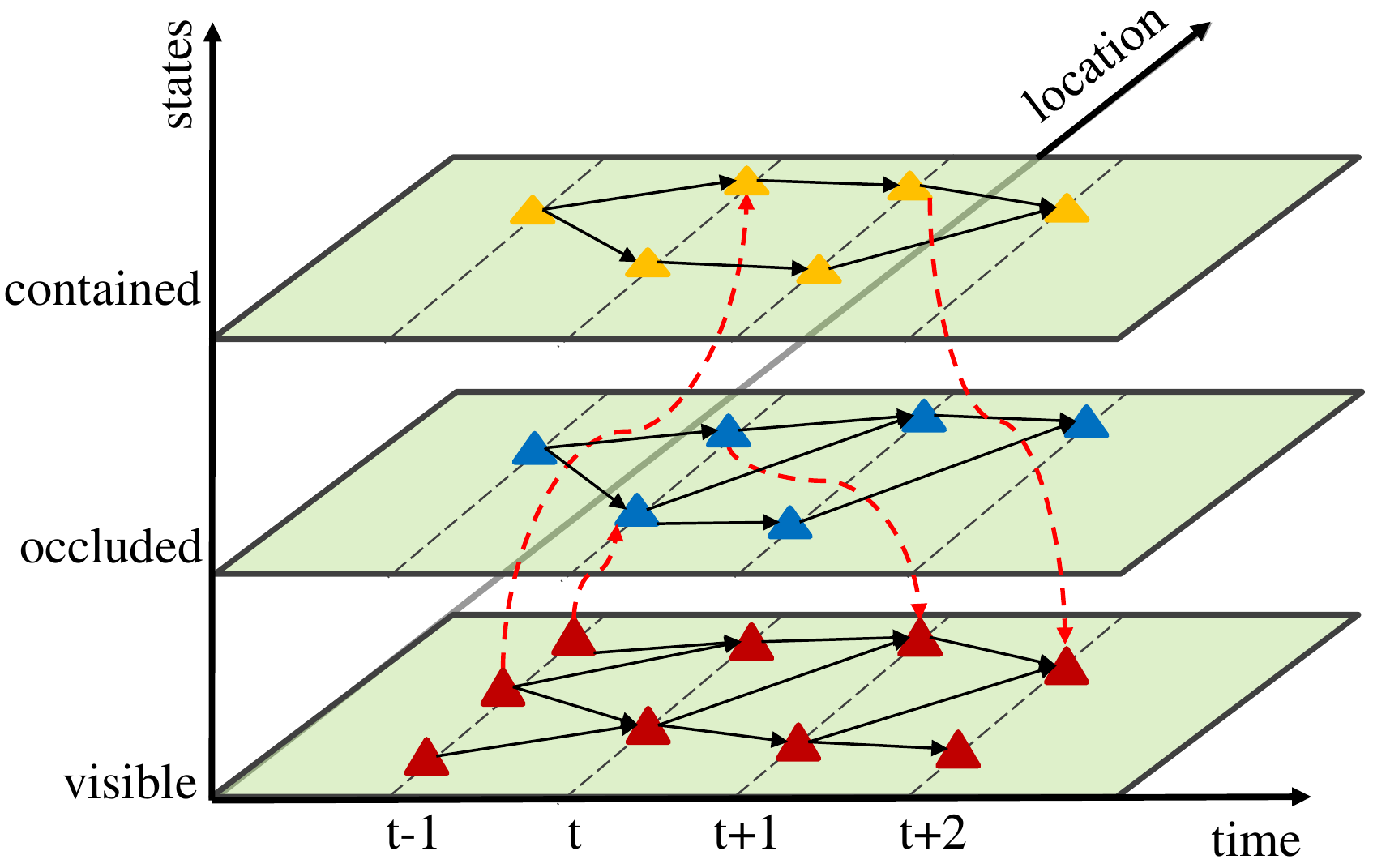}
\caption{\textbf{Transition graph utilized to formulate the integer linear programming.} Each node $m$ has its location $l_m$, state $s_m$, and time instant $t_m$. Black solid arrows indicate the possible transitions in the same state. Red dashed arrows indicate the possible transitions between different states.}
\label{fig:state}
\vspace{-10pt}
\end{figure}

We cast the intractable optimization of Eqn.~\eqref{equation:score} as an Integer Linear Formulation (ILF) in order to derive a scalable and efficient inference algorithm.
We use $V$ to denote the locations of vehicles, and $E$ to denote the edges between all possible pairs of nodes, whose time is consecutive and locations are close. The whole transition graph $G = (V, E)$ is shown as Fig.~\ref{fig:state}. Then the energy function Eqn.~\eqref{equation:score} can be re-written as:
\begin{equation}\small\label{equation:ILF} \begin{aligned}
    f^* & = \arg\max_f\, \sum_{mn\in E_o}c_{mn}f_{mn}, \\
    c_{mn} & = - \Phi(l_n,l_m,s_m) - \Psi(s_n,s_m,a_m) - \Upsilon(l_m,\phi_m,s_m)  \\
           & \quad\, - \Gamma(l_m,\phi_m,a_m), \\
    s.t. & \ \ \  f_{mn}\in \{0 ,1\}, \; \sum_{m}f_{mn} \leq 1, \; \sum_{m}f_{mn} = \sum_{k}f_{nk},
\end{aligned} \end{equation}
where $f_{mn}$ is the number of object moving from node $V_m$ to node $V_n$, $c_{mn}$ is the corresponding cost.

Since the subject of interest can only enter a nearby container (e.g., vehicle), to discover the optimal causal parse graph, we need to jointly track the container and the subject of interest. Similar to Eqn.~\eqref{equation:ILF}, the energy function of container is as follows:
\begin{equation}\small\label{equation:container} \begin{aligned}
    & g^* = \arg\max_g\, \sum_{mn\in E_c} d_{mn}\, g_{mn}, \\
    & d_{mn} = h_c(l_m,\phi_m) - 1, \\
    s.t. \ \ \ & g_{mn} \in \{0 ,1\}, \; \sum_{m} g_{mn} \leq 1, \; \sum_{m}g_{mn} = \sum_{k}g_{nk},
\end{aligned} \end{equation}
where $h_c(l_m,\phi_m)$ is the container detection score at location $l_m$. Then we add the contained constrains as:
\begin{equation}\label{equation:combine} \begin{aligned}
    & \sum_{mn\in E_c}g_{mn}\geq \sum_{ij\in E_o}f_{ij}, \\
    & s.t. \;\;  t_n = t_j,\; \| l_n - l_j \|_2 < \tau_c,
\end{aligned} \end{equation}
where $\tau_c$ is the distance threshold. Finally, we combine Eqn.~\eqref{equation:ILF}-\eqref{equation:combine} to obtain objective function for our model:
\begin{equation}\small\begin{aligned} \label{equation:MIP}
    f^*, g^* &= \max_{f,g}\; \sum_{mn\in E_o}c_{mn}f_{mn} + \sum_{ij\in E_c} d_{mn}\,g_{mn}, \\
    s.t.\;\; & f_{mn}\in \{0 ,1\}, \; \sum_{m}f_{mn} \leq 1, \; \sum_{m}f_{mn} = \sum_{k}f_{nk}, \\
    	     & g_{mn}\in \{0 ,1\}, \; \sum_{i}g_{mn} \leq 1, \; \sum_{i}g_{mn} = \sum_{k}g_{nk}, \\
    	     & t_n = t_j,\; \| l_n - l_j \|_2 < \tau_c.
\end{aligned} \end{equation}
The re-formulated graph still follows a directed acyclic graph (DAG). Thus we can adopt the Dynamic Programming technique to efficiently search for the optimal solution, as illustrated in the Fig.~\ref{fig:state}.

\vspace{-2mm}
\section{Experiments} \label{sec:exp}
\vspace{-1mm}

We apply the proposed method on two tracking interacting objects datasets and evaluate the improvement in visual tracking brought by the outcomes of visibility status reasoning.

\vspace{-2mm}
\subsection{Implementation Details}
\vspace{-1mm}

We first utilize the Faster R-CNN models~\cite{ren2015faster} trained on the MS COCO dataset to detect involved agents (e.g., person and suitcase). The used network is the VGG-16 net, with score threshold 0.4 and NMS threshold 0.3. The tracklets similarity threshold $\tau_\varsigma$ is set as 0.8. The contained distance threshold $\tau_c$ is set as the width of container $3$ meters. The maximum number of contained objects in a container is set to $5$. For appearance descriptors $\phi$, we employ the dense sampling ColorNames descriptor~\cite{zheng2015scalable}, which applies square root operator~\cite{arandjelovic2012three} and Bag-of-word encoding to the original ColorNames descriptors. For human skeleton estimation, we use the public implementation of~\cite{wei2016convolutional}. For vehicle detection and semantic part status estimation, we use the implementation provided by~\cite{li2016recognizing} with default parameters mentioned in their paper.

We adopt the widely used CLEAR metrics~\cite{kasturi2009framework} to measure the performances of tracking methods. It includes four metrics, i.e., Multiple Object Detection Accuracy (MODA), Detection Precision (MODP), Multiple Object Tracking Accuracy (MOTA) and Tracking Precision (MOTP), which take into account three kinds of tracking errors: false positives, false negatives and identity switches. We also report the number of false positives (FP), false negatives (FN), identity switches (IDS) and fragments (Frag). A higher value means better for MODA, MODP, MOTA and MOTP, while a lower value means better for FP, FN, IDS and Frag. If the Intersection-over-Union (IoU) ratio of tracking results to groundtruth is above 0.5, we accept the tracking result as a correct hit.

\vspace{-2mm}
\subsection{Datasets}
\vspace{-1mm}

\begin{table*}[ptb]
\centering
\begin{threeparttable}
\resizebox{\linewidth}{!}{
\renewcommand\arraystretch{1.05}
\setlength\tabcolsep{2pt}
\begin{tabular}{l||l||c|c|c|c|c|c|c|c|c|c|c|c}
\hline\thickhline
People-Car & Metric    &Our-full &Our-1 &Our-2 &POM~\cite{fleuret2008multicamera} &SSP~\cite{pirsiavash2011globally} &LP2D~\cite{leal2014learning} &LP3D~\cite{leal2014learning} &KSP-fixed~\cite{berclaz2011multiple}
&KSP-free~\cite{berclaz2011multiple} &KSP-seq~\cite{berclaz2011multiple}
&TIF-LP~\cite{TrackInteract16} &TIF-MIP~\cite{TrackInteract16} \\
\hline
\multirow{4}{*}{Seq.0}
& FP $\downarrow$ &0.17&0.34&0.20&0.06 &0.04 &0.05 &0.05 &0.46 &0.10 &0.46 &0.07 &0.07\\
& FN $\downarrow$ &0.08&0.53&0.12&0.47 &0.76 &0.48 &0.53 &0.61 &0.41 &0.61 &0.25 &0.25\\
& IDS $\downarrow$ &0.05&0.07&0.05&- &0.04 &0.06 &0.06 &0.07 &0.07 &0.07 &0.04 &0.04\\
& MODA $\uparrow$ &\textbf{0.71}&0.27&0.63 &0.47 &0.20 &0.47 &0.42 &-0.07 &0.49 &-0.07 &0.67 &0.67\\
\hline
\multirow{4}{*}{Seq.1}
& FP $\downarrow$ &0.21&0.70&0.28&0.98 &0.75 &0.77 &0.75 &0.77 &0.71 &0.75 &0.17 &0.17\\
& FN $\downarrow$ &0.12&0.26&0.14&0.23 &0.25 &0.21 &0.25 &0.25 &0.25 &0.25 &0.25 &0.25\\
& IDS $\downarrow$ &0.04&0.13&0.04&- &0.12 &0.17 &0.21 &0.06 &0.12 &0.15 &0.04 &0.04\\
& MODA $\uparrow$ &\textbf{0.62}&0.09&0.54&-0.21 &0.00 &0.02 &0.00 &-0.02 &0.04 &0.00 &0.58 &0.58\\
\hline
\multirow{4}{*}{Seq.2}
& FP $\downarrow$ &0.03&0.05&0.04&0.03 &0.00 &0.03 &0.00 &0.05 &0.00 &0.05 &0.03 &0.03\\
& FN $\downarrow$ &0.28&0.58&0.32&0.47 &0.59 &0.62 &0.58 &0.72 &0.59 &0.72 &0.47 &0.47\\
& IDS $\downarrow$ &0.01&0.03&0.02&- &0.01 &0.02 &0.01 &0.03 &0.01 &0.03 &0.01 &0.01\\
& MODA $\uparrow$ &\textbf{0.57}&0.39&0.48&0.50 &0.41 &0.35 &0.42 &0.23 &0.41 &0.23 &0.50 &0.50\\
\hline
\multirow{3}{*}{Seq.3}
& FP $\downarrow$ &0.18&0.39&0.21&0.59 &0.35 &0.43 &0.27 &0.46 &0.43 &0.43 &0.14 &0.14\\
& FN $\downarrow$ &0.07&0.32&0.10&0.17 &0.31 &0.23 &0.40 &0.19 &0.23 &0.19 &0.21 &0.21\\
& IDS $\downarrow$ &0.06&0.26&0.06&- &0.27 &0.34 &0.33 &0.19 &0.25 &0.21 &0.07 &0.05\\
& MODA $\uparrow$ &\textbf{0.68}&0.35&0.62&0.24 &0.34 &0.34 &0.33 &0.35 &0.34 &0.38 &0.65 &0.65\\
\hline
\multirow{3}{*}{Seq.4}
& FP $\downarrow$ &0.16&0.27&0.18&0.40 &0.19 &0.26 &0.13 &0.32 &0.25 &0.31 &0.08 &0.07\\
& FN $\downarrow$ &0.10&0.18&0.13&0.15 &0.19 &0.16 &0.18 &0.17 &0.17 &0.16 &0.16 &0.15\\
& IDS $\downarrow$ &0.05&0.15&0.05&- &0.14 &0.13 &0.15 &0.12 &0.12 &0.11 &0.04 &0.04\\
& MODA $\uparrow$ &\textbf{0.82}&0.59&0.73&0.45 &0.62 &0.58 &0.69 &0.51 &0.58 &0.53 &0.76 &0.78\\
\hline
\end{tabular}
}
\end{threeparttable}
\vspace*{0pt}
\caption{\textbf{Quantitative results and comparisons} of false positive (FP) rate, false negative (FN) rate and identity switches (IDS) rate \textbf{on People-Car Dataset}. The best scores are
marked in \textbf{bold}.}
\label{tab:interaction}
\vspace{-10pt}
\end{table*}

\textbf{People-Car dataset}~\cite{TrackInteract14}\footnote{This dataset is available at \url{cvlab.epfl.ch/research/surv/interacting-objects}}. This dataset consists of $5$ groups of synchronized sequences on a parking lot, recorded from two calibrated bird-view cameras, with length of $300\sim5100$ frames. In this dataset, there are many instances of people getting in and out of cars. This dataset is challenging for the frequent interactions, light variation and low object resolution.

\textbf{Tracking Interacting Objects (TIO) dataset}. For current popular multiple object tracking datasets (e.g., PETS09~\cite{Ferryman2009}, KITTI dataset~\cite{geiger2012we}), most tracked objects are pedestrian and no evident interaction visibility fluent changes. Thus we collect two new scenarios with typical human-object interactions: person, suitcase, and vehicle on several places.

\textit{Plaza}. We capture $22$ video sequences in a plaza that describe people walking around, getting in/out vehicles.

\textit{ParkingLot}. We capture $15$ video sequences in a parking lot that shows vehicles entering/exiting the parking lot, people getting in/out vehicles, people interacting with trunk/suitcase.

All video sequences are captured by a GoPro camera, with frame rate 30fps and resolution $1920\times1080$. We use the standard chessboard and Matlab camera calibration toolbox to obtain camera parameters. The total number of frames of TIO dataset is more than 30K. There exist severe occlusions and large scale changes, making this dataset very challenging for traditional tracking methods.

\begin{table}[ptb]
\begin{center}
\renewcommand\arraystretch{1.05}
\setlength\tabcolsep{4pt}
\resizebox{\linewidth}{!}{
\begin{tabular}{l||c|c|c|c|c|c}
\hline\thickhline
Plaza      &MOTA $\uparrow$&MOTP $\uparrow$&FP $\downarrow$&FN $\downarrow$&IDS $\downarrow$&Frag $\downarrow$ \\
\hline
Our-full    &\textbf{46.0}\%&76.4\%&99&501&5 &8\\
Our-1       &31.9\%&75.1\%&40&643&29&36\\
Our-2       &32.5\%&75.3\%&75&605&25&30\\
MHT\_D~\cite{kim2015multiple}      &34.3\%&73.8\%&56&661&15&18\\
MDP~\cite{xiang2015learning}         &32.9\%&73.2\%&24&656&9&7\\
DCEM~\cite{milan2016multi}        &32.3\%&\textbf{76.5}\%&2&  675&   2&    2\\
SSP~\cite{pirsiavash2011globally}        &31.7\%&72.1\%&19&678&21&25\\
DCO~\cite{andriyenko2012discrete}         &29.5\%&76.4\%&22&673&6&2\\
JPDA\_m~\cite{hamid2015joint}     &13.5\%&72.2\%&163&673&6&3\\
\hline\cline{1-7}
ParkingLot   &MOTA $\uparrow$&MOTP $\uparrow$&FP $\downarrow$&FN $\downarrow$&IDS $\downarrow$&Frag $\downarrow$ \\
\hline
Our-full       &\textbf{38.6\%}    &\textbf{78.6\%}    &418    &1954    &6    &5\\
Our-1          &28.7\%    &78.4\%    &451    &2269    &15    &17\\
Our-2          &28.9\%    &78.4\%    &544    &2203    &14    &16\\
MDP~\cite{xiang2015learning}            &30.1\%    &76.4\%    &397    &2296    &26    &22\\
DCEM~\cite{milan2016multi}          &29.4\%    &77.5\%    &383    &2346    &16    &15\\
SSP~\cite{pirsiavash2011globally}            &28.9\%    &75.0\%    &416    &2337    &12    &14\\
MHT\_D~\cite{kim2015multiple}         &25.6\%    &75.7\%    &720    &2170    &15    &12\\
DCO~\cite{andriyenko2012discrete}            &24.3\%    &78.1\%    &536    &2367    &38    &10\\
JPDA\_m~\cite{hamid2015joint}        &12.3\%    &74.2\%    &1173   &2263    &28    &17\\
\hline    	    	
\end{tabular}}
\end{center}
\caption{\textbf{Quantitative results and comparisons} of false positive (FP), false negative (FN), identity switches (IDS), and  fragments (Frag) on \textbf{TIO dataset}. The best scores are marked in \textbf{bold}.}
\label{tab:plaza}
\vspace{-10pt}
\end{table}

Beside the above testing data, we collect another set of video clips for training. To avoid over-fitting, we set up different camera positions, different people and vehicles from the testing settings. The training data consists of $380$ video clips covering 9 events: \emph{walking, opening vehicle door, entering vehicle, exiting vehicle, closing vehicle door, opening vehicle trunk, loading baggage, unloading baggage, closing vehicle trunk}. Each action category contains 42 video clips on average.

Both the datasets and short video clips are annotated with bounding boxes for people, suitcases, vehicles, and visibility fluents of people and suitcases. The types of status are ``visible'', ``occluded'', and ``contained''. We utilize VATIC ~\cite{vondrick2013efficiently} to annotate the videos.

\vspace{-2mm}
\subsection{Results and Comparisons}
\vspace{-1mm}

For People-Car dataset, we compare our proposed method with $5$ baseline methods and their variants: successive shortest path algorithm (SSP)~\cite{pirsiavash2011globally}, K-Shortest Paths Algorithm
(KSP-fixed, KSP-free, KSP-seq)~\cite{berclaz2011multiple}, Probability Occupancy Map (POM)~\cite{fleuret2008multicamera}, Linear Programming (LP2D, LP3D)~\cite{leal2014learning}, and Tracklet-Based Intertwined Flows (TIF-IP, TIF-MIP)~\cite{TrackInteract16}. We refer the reader to~\cite{TrackInteract16} for more details about the method variants. The quantitative results are reported in Table~\ref{tab:interaction}. From the results, we can observe that the proposed method obtains better performance than the baseline methods.

For TIO dataset, we compare the proposed method with $6$ state-of-the-arts: successive shortest path algorithm (SSP)~\cite{pirsiavash2011globally}, multiple hypothesis tracking with distinctive appearance model (MHT\_D)~\cite{kim2015multiple}, Markov Decision Processes with Reinforcement Learning (MDP)~\cite{xiang2015learning}, Discrete-Continuous Energy Minimization (DCEM)~\cite{milan2016multi}, Discrete-continuous optimization (DCO) ~\cite{andriyenko2012discrete} and Joint Probabilistic Data Association (JPDA\_m)~\cite{hamid2015joint}. We use the public implementations of these methods.

\begin{figure*}[ptb]
\centering	
\includegraphics[width=0.99\linewidth]{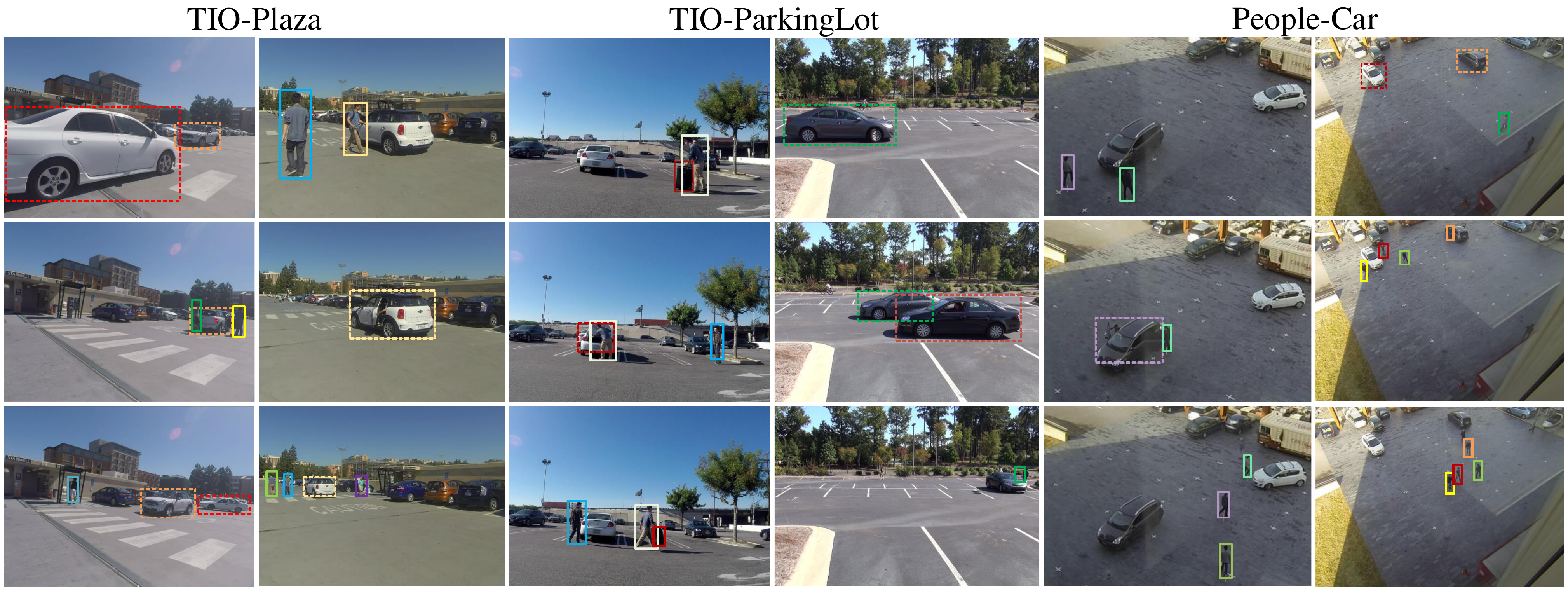}
\caption{\textbf{Sampled qualitative results of our proposed method on TIO dataset and People-Car dataset.} Each color represents an object. The solid bounding box means the visible object. The dash bounding box denotes the object is contained by other scene entities. Best viewed in color and zoom in.}
\label{fig:result_table}
\vspace{-10pt}
\end{figure*}

We report quantitative results and comparisons in Table~\ref{tab:plaza} for TIO dataset. From the results, we can observe that our method obtains superior performance to the other methods on most metrics. It validates that the proposed method can not only track visible objects correctly, but also reason locations for occluded or contained objects. The alternative methods do not work well mainly due to lack of the ability to track objects under long-term occlusion or containment in other objects.

\begin{figure}[htb]
\centering
\includegraphics[width=\linewidth]{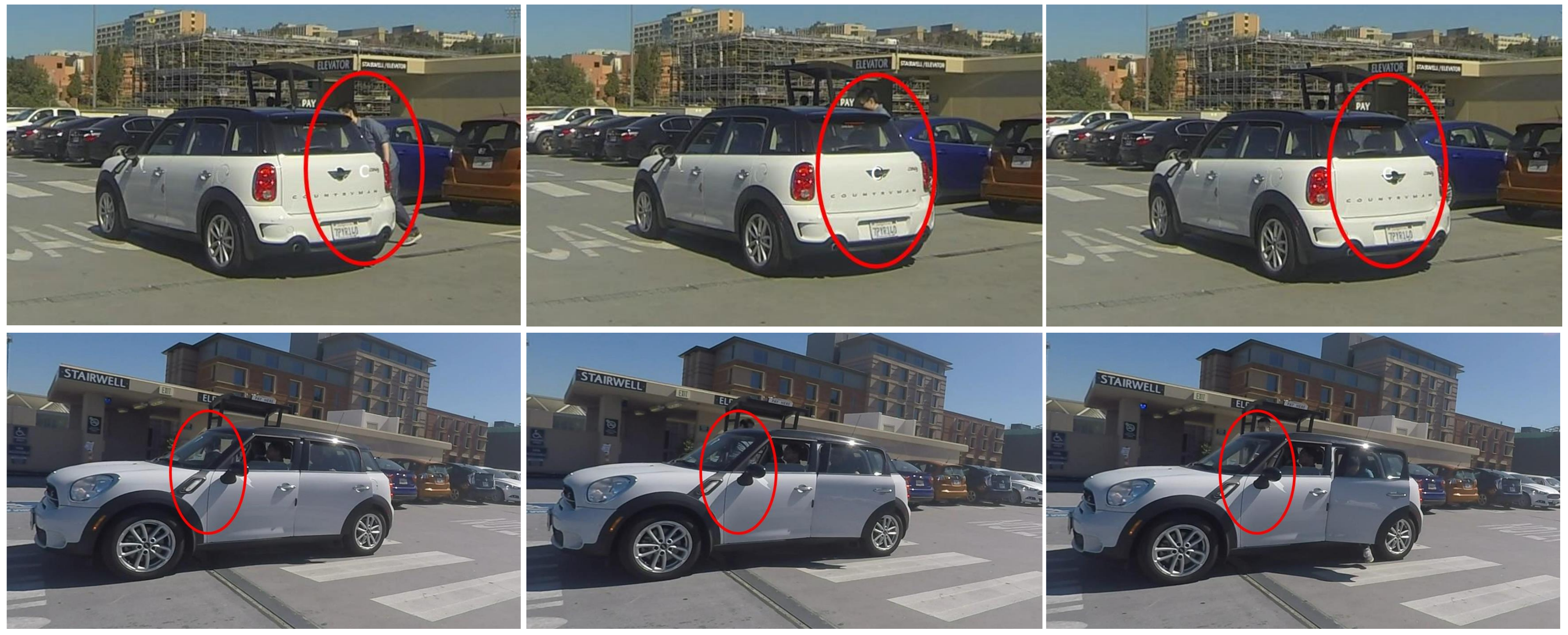}
\caption{\textbf{Sampled failure cases}. When people stay behind vehicles, it is hard to determine whether or not they are interacting with the vehicle, e.g., entering, exiting.}
\label{fig:fail_case}
\vspace{-5pt}
\end{figure}

\begin{figure}[htb!]
\centering
\includegraphics[width=0.9\linewidth]{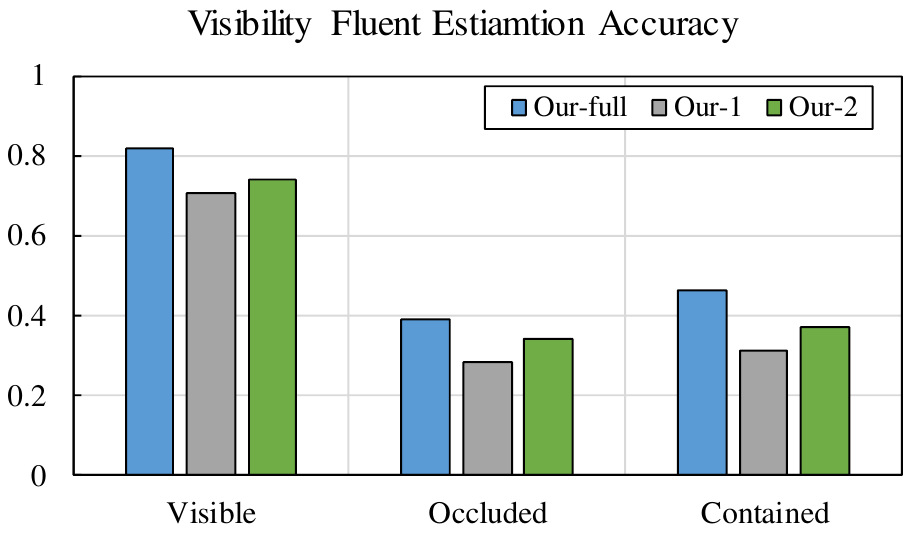}
\caption{\textbf{Visibility fluent estimation} results on TIO dataset.}
\label{fig:state_est}
\vspace{-10pt}
\end{figure}

We set up three baselines to analyze the effectiveness of different components in the proposed method:

\noindent{\small\textbullet}~\textbf{Our-1}: no likelihood term and only prior term is used.

\noindent{\small\textbullet}~\textbf{Our-2}: only human data-likelihood term and prior term are used.

\noindent{\small\textbullet}~\textbf{Our-full}: all terms are used, including prior terms, human and vehicle data-likelihood terms.

Based on comparisons of Our-1, Our-2 and Our-full, we can also conclude that each type of fluent plays its role in improving the final tracking results. Some qualitative results are displayed in Fig.~\ref{fig:result_table}.

We further report fluent estimation results on TIO-Plaza sequences and TIO-ParkingLot sequences in Fig.~\ref{fig:state_est}. From the results, we can see that our method can successfully reason the visibility status of subjects. Note that the precision of containment estimation is not high, since some people get in/out the vehicle from the opposite side towards the camera, as shown in Fig.~\ref{fig:fail_case}. Under such situation, there are barely any image evidence to reason the object status and multi-view setting might be a better way to reduce the ambiguities.

\vspace{-2mm}
\section{Conclusion}
\vspace{-1mm}

In this paper, we propose a Causal And-Or Graph (C-AOG) model to represent the causal-effect relations between object visibility fluents and various human interactions. By jointly modeling short-term occlusions and long-term occlusions, our method can explicitly reason the visibility of subjects as well as their locations in the videos. Our method clearly outperforms the alternative methods in complicated scenarios with frequent object interactions. In this work, we focus on the human-interactions as a running-case of the proposed technique, and we will explore the extension of our method to other types of objects (e.g., animal, drones) in the future.

{
\small
\bibliographystyle{ieee}
\bibliography{st_reasoning}
}

\end{document}